\documentclass{article}
\usepackage[utf8]{inputenc}

\usepackage{pgfplots}       

\pgfplotsset{compat=1.18}

\usepackage[pdfencoding=auto]{hyperref}

\usepackage{bm}

\usepackage{geometry}
\usepackage{graphicx}

\usepackage[citestyle=authoryear, backend=bibtex, sorting=nyt, natbib=true, url=false, doi=false]{biblatex}
\usepackage{amsmath}
\usepackage{amssymb} 
\usepackage{nicefrac}  
\usepackage{float} 
\usepackage{amsthm} 

\usepackage{hyperref} 

\usepackage{xcolor} 

\usepackage{thmtools} 

\declaretheorem[name=Hypothesis, preheadhook={}]{hyp} 

\usepackage{cleveref} 
\crefname{hyp}{hypothesis}{hypotheses} 
\Crefname{hyp}{Hypothesis}{Hypotheses} 

\usepackage{ragged2e} 

\usepackage{booktabs} 
\usepackage{multirow} 

\usepackage{algorithm}
\usepackage{algpseudocode}

\algnewcommand{\IfThen}[2]{
	\State \algorithmicif\ #1\ \algorithmicthen\ #2}
\algnewcommand{\LineComment}[1]{\State \(\triangleright\) #1}

\algnewcommand{\InlineFor}[2]{
	\State \algorithmicfor\ #1\ \algorithmicendfor\ #2}

 \makeatletter
\AfterEndEnvironment{algorithm}{\let\@algcomment\relax}
\AtEndEnvironment{algorithm}{\kern2pt\hrule\relax\vskip3pt\@algcomment}
\let\@algcomment\relax
\newcommand\algcomment[1]{\def\@algcomment{\footnotesize#1}}

\renewcommand\fs@ruled{\def\@fs@cfont{\bfseries}\let\@fs@capt\floatc@ruled
  \def\@fs@pre{\hrule height.8pt depth0pt \kern2pt}%
  \def\@fs@post{}%
  \def\@fs@mid{\kern2pt\hrule\kern2pt}%
  \let\@fs@iftopcapt\iftrue}
\makeatother

\usepackage{booktabs}

\usepackage[parfill]{parskip} 

\usepackage{authblk}

\usepackage{algorithm}
\usepackage{algpseudocode}

\providecommand{\keywords}[1]
{
  \small	
  \textbf{\textit{Keywords---}} #1
}

\title{Hybrid Node-Destroyer Model with Large Neighborhood Search for Solving the Capacitated Vehicle Routing Problem}

\author[1]{Bachtiar Herdianto}
\author[1]{Romain Billot}
\author[2]{Flavien Lucas}
\author[3]{Marc Sevaux}
\author[4]{Daniele Vigo}

\date{\color{blue} 
      \nolinkurl{bachtiar.herdianto@imt-atlantique.fr}$^{*1}$\\
      \nolinkurl{romain.billot@imt-atlantique.fr}$^1$\\
      \nolinkurl{flavien.lucas@imt-nord-europe.fr}$^2$\\
      \nolinkurl{marc.sevaux@univ-ubs.fr}$^3$\\
      \nolinkurl{daniele.vigo@unibo.it}$^4$}

\affil[1]{IMT Atlantique, Lab-STICC (UMR 6285, CNRS), Brest, France}
\affil[2]{IMT Nord Europe, CERI Systèmes Numériques, Douai, France}
\affil[3]{Université Bretagne Sud, Lab-STICC (UMR 6285, CNRS), Lorient, France}
\affil[4]{DEI "G. Marconi", University of Bologna, Bologna, Italy}

\begin{document}

\maketitle

\begin{abstract}
In this research, we propose an iterative learning hybrid optimization solver developed to strengthen the performance of metaheuristic algorithms in solving the Capacitated Vehicle Routing Problem (CVRP). The iterative hybrid mechanism integrates the proposed Node-Destroyer Model, a machine learning hybrid model that utilized Graph Neural Networks (GNNs) such identifies and selects customer nodes to guide the Large Neighborhood Search (LNS) operator within the metaheuristic optimization frameworks. This model leverages the structural properties of the problem and solution that can be represented as a graph, to guide strategic selections concerning node removal. The proposed approach reduces operational complexity and scales down the search space involved in the optimization process. The hybrid approach is applied specifically to the CVRP and does not require retraining across problem instances of different sizes. The proposed hybrid mechanism is able to improve the performance of baseline metaheuristic algorithms. Our approach not only enhances the solution quality for standard CVRP benchmarks but also proves scalability on very large-scale instances with up to $30,000$ customer nodes. Experimental evaluations on benchmark datasets show that the proposed hybrid mechanism is capable of improving different baseline algorithms, achieving better quality of solutions under similar settings.

\end{abstract}


\section{Introduction}
\label{sec:introduction}
    Routing is an important part of logistics and supply chains, involving the transportation of goods from one location to another, directly influencing pricing structures in the market \citep{Simchi-Levi2000-fd,arnold2019makes}. A challenge arises whenever delivery costs increase, which can affect pricing structures. As a result, optimizing delivery routes has become important. One of the most extensively studied problems in this context is the Capacitated Vehicle Routing Problem (CVRP), continues to pose significant challenges in both academic research and industrial practice \citep{laporte2009fifty,arnold2019knowledge,accorsi2021fast,leng2021distribution,yin2022multiobjective}. Most approaches to solving the CVRP have relied on heuristics and metaheuristics, typically grounded in human intuition \citep{arnold2019makes,arnold2019knowledge}. However, recent years have witnessed growing interest in applying machine learning (ML) to enhance optimization performance \citep{Hottung_Andr__2020,bengio2021machine,ke2022deep}. The integration of ML with optimization can be categorized into three main strategies: (1) end-to-end learning, (2) learning based on problem properties, and (3) learning from repeated decisions \citep{bengio2021machine}. The third approach, in particular, enables the development of in-loop, ML-assisted optimization algorithms that can dynamically adjust their behavior. This approach allows the algorithm to learn from its own decisions, thereby improving performance over time. 
    
    One example of this mechanism is Neural LNS (NLNS) \citep{Hottung_Andr__2020, HOTTUNG2022103786}, which aims to control the LNS using a sequence model \citep{sutskever2014sequence}. However, these learned LNS approaches have been applied only to small CVRP instances, where existing state-of-the-art metaheuristics already perform very well. Beyond the need for improved scalability, these approaches also require statistical evidence to validate their effectiveness. Meanwhile, a graph-based end-to-end solvers \citep{joshi2019efficient} leverage GNN to generate approximate solutions directly from problem instances. Building on this idea, heatmaps can be employed to guide local search operators \citep{hudson2022graph}, Monte Carlo Tree Search (MCTS) \citep{xia2024position}, and even simple Dynamic Programming \citep{kool2022deep}. The heatmap-based approach, initially proposed for small-scale problems \citep{joshi2019efficient}, has been generalized to handle large-scale Traveling Salesperson Problem (TSP) instances \citep{xin2021neurolkh,ye2023deepaco,kim2025ant}. It has also been extended to more complex variant of problems, such as the TSP with Time Windows (TSPTW), and CVRP \citep{xin2021neurolkh,kool2022deep,ye2023deepaco,kim2025ant}.
    
    Built on these advances, this research aims to solve the CVRP by developing a hybrid \textit{learning-from-repeated-decisions mechanism} that leverages GNNs as the base model. Specifically, we extend the hybrid GNN-based optimization solver to guide the LNS operator by focusing on the selection of customer nodes, thereby steering the search process toward more promising solution spaces.

\begin{figure}[H]
    \centering 
    \includegraphics[width=1\textwidth]{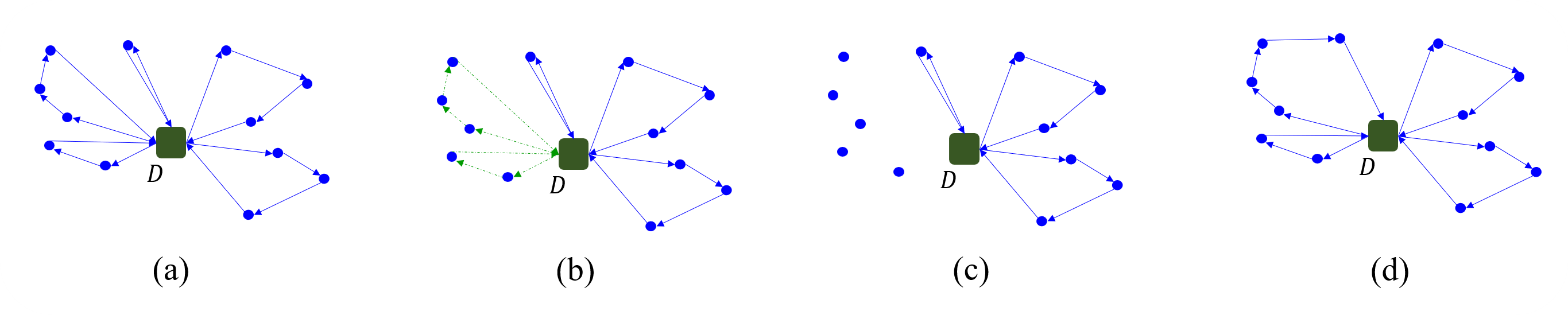}
    \caption{Overview of Large Neighborhood Search.} \label{fig:large-neighborhood-search} 
\end{figure}

\section{Related Works}
\label{sec:related-works}
    The CVRP remains an essential optimization problem due to its practical relevance in logistics and transportation \citep{leng2021distribution,yin2022multiobjective}. Despite the high application, current exact methods are limited in size and far below real-world cases \citep{Prodhon2016}. Due to their effectiveness at large scales, metaheuristics have become a widely favored approach, with vehicle routing standing out as a key application \citep{Prodhon2016,leng2021distribution}. While small local search, such as 2-opt \citep{croes1958method}, 2-opt$*$ \citep{Potvin01121995}, CROSS-exchange \citep{taillard1997tabu}, etc., offer efficient computational complexity by exploring small neighborhoods, their limited size often restricts the search to local optimal \citep{PISINGER20072403}.

    \subsection{Large Neighborhood Search (LNS)}
    \label{subsec:lns}
        To overcome the limitation of small local search, LNS \citep{PISINGER20072403} was proposed as an extension of the Large Neighborhood Search (LNS). LNS follows a continual relaxation and re-optimization process, where, particularly in VRPs, selected customer nodes are removed from the routing plan and reinserted during re-optimization. As shown in \Cref{fig:large-neighborhood-search}, in (a), an initial solution is shown, where routes emanate from a depot, and each customer is visited exactly once. Thus, in (b) represents the \textit{destroy} phase of LNS, where a subset of customers is partially removed from the current solution to dismantle its structure. In (c), the removed customers are displayed separately from the remaining routes and are awaiting reinsertion. This phase highlights the neighborhood that will be explored for potential improvements. Finally, (d) illustrates the \textit{repair} phase, where the removed customers are reinserted into the solution. This iterative destroy-and-repair approach enables LNS to explore a larger solution space. The Adaptive Large Neighborhood Search (ALNS) extends LNS by employing multiple destroy and repair operators, which are adaptively selected based on their past performance \citep{ropke2006adaptive,masson2013adaptive,christiaens2020slack,arnold2021pils,accorsi2021fast}. The effectiveness of LNS depends on several configurable components, each of which can be implemented in various ways to suit different problem characteristics \citep{VOIGT2025357}. The \textit{search space} in LNS can be restricted to only feasible solutions, ensuring all explored candidates satisfy problem constraints. Alternatively, infeasible solutions may be allowed, with violations penalized through a generalized cost function to guide the search toward feasibility \citep{arnold2021pils}. The \textit{starting solution}, a critical aspect that influences the early performance of the algorithm, can be generated using simple insertion operators or more sophisticated construction heuristics that build feasible solutions, for example, the Clarke and Wright saving algorithm improved by a particular local search operator \citep{christiaens2020slack,accorsi2021fast}. \textit{Operators} are the important components of the Large Neighborhood Search, and their variety affects performance. This includes the number and type of removal operators, which determine how parts of the solution are disrupted, and insertion operators, which rebuild the partial solution. \textit{Local search} is another component in LNS, used to refine solutions further. It may be applied conditionally or universally (after every iteration). Finally, the \textit{acceptance criterion} is a key component in LNS, with common strategies such as simulated annealing, which probabilistically accepts worse solutions to escape local optima. Recent work, like the Slack Induction by String Removals (SISRs) metaheuristic \citep{christiaens2020slack}, proposes executing a large number of small and fast iterations to offset the limited local search capabilities of LNS. However, the overall effectiveness of LNS remains highly sensitive to the choice and combination of destroy and repair operators \citep{accorsi2021fast,VOIGT2025357}.

    \subsection{Hybrid GNNs and Optimization Algorithm}
    \label{subsec:hybrid-gnn-optimization}
        GNNs have been successfully applied across a wide range of domains \citep{joshi2019efficient,xin2021neurolkh,hudson2022graph,rahmani2023graph,ye2023deepaco,kim2024neural,ye2024glop,kim2025ant,ouyang2025learningsegmentvehiclerouting}. Their ability to process and learn from graph-structured data is particularly important in combinatorial optimization, where many problems can naturally be formulated as graphs \citep{bengio2021machine}. Recent studies have explored hybrid techniques that integrate GNNs with optimization algorithms for solving routing problems. For instance, NeuroLKH \citep{xin2021neurolkh} combines the graph ConvNet \citep{joshi2019efficient} with LKH \citep{helsgaun2017extension}, where the graph ConvNet assigns edge scores and node penalties to guide the optimization. Similarly, the GNN-GLS \citep{hudson2022graph} employs a graph ConvNet to predict regret values for each edge, guiding the Guided Local Search (GLS) \citep{arnold2019knowledge} on which edges to penalize. NeuralSEP \citep{kim2024neural} uses a GNN \citep{gilmer2017neural} to approximate the RCI separation problem, integrating the learned model into an exact, cutting-plane method. In the context of metaheuristics, DeepACO \citep{ye2023deepaco} leverages GNN \citep{joshi2019efficient,NEURIPS2022_a3a7387e} to predict heatmaps that bias solution construction and guide the probabilistic search process of Ant Colony Optimization (ACO) \citep{dorigo2007ant}. This was later extended by the Generative Flow Ant Colony Sampler (GFACS) \citep{kim2025ant}, which incorporates Generative Flow Networks (GFlowNets) \citep{bengio2021flow} to learn reward-proportional sampling distributions for guiding ant sampling and pheromone updates. Efforts have also been made to combine graph learning with sequential learning models. For example, the Global and Local Optimization Policies GLOP \citep{ye2024glop} adopts a decomposition strategy, where graph learning first analyzes the problem space to generate partitions, which are then refined by sequential learning. Similarly, in the context of decomposition, Learning to Segment (L2Seg) \citep{ouyang2025learningsegmentvehiclerouting} introduces a segmentation mechanism that identifies stable parts of a CVRP solution, grouping them into fixed segments for the next search phase, thereby allowing LKH to refine only the remaining parts. In the context of LNS, Adaptive Dynamic Neighborhood Search (ADNS) \citep{WANG2025113280} utilizes a GNN \citep{veličković2018graph} to embed the current solution and adaptively determine neighborhood structures for LNS operations. However, challenges remain in terms of scalability and generalizability. Many of the aforementioned methods require fine-tuning for each problem variant. Furthermore, the quality of solutions often lags behind that of established state-of-the-art solvers, indicating room for further improvement.
        
    \subsection{Research Questions and Contributions}
    \label{subsec:ch5-research-question-and-contributions}
        The LNS extends Local (Neighborhood) Search by dynamically employing a diverse set of destroy and repair operators. This enables the exploration of broader solution neighborhoods, thereby enhancing the ability of the algorithm to escape local optima and converge toward higher-quality solutions \citep{ropke2006adaptive}. However, the performance of LNS relies on how these operators are selected during the search process \citep{VOIGT2025357}. Recent advances in hybrid graph learning based optimization have shown the potential of GNNs to learn structural patterns in combinatorial optimization \citep{joshi2019efficient,Joshi2022,kool2022deep,rahmani2023graph,xu2020neural}. Motivated by this, we aims to investigate the hybridization of graph-based learning into the LNS framework for solving the CVRP. In summary, we pose the following questions:
        
        \begin{enumerate}
            \item How can we develop a selector model that identifies customer nodes whose removal leads to more effective processes in LNS?
            \item How can we design a hybrid mechanism that leverages this graph-based model to enhance the performance and adaptability of LNS in solving the CVRP?
        \end{enumerate}
        
        To explore the integration of GNNs into the metaheuristic framework, firstly, we aim to develop a selector model capable of identifying customer nodes whose removal is most likely to yield improvements. This model leverages structural information from the problem instance and the quality of solution. Building on this, we design a hybrid mechanism that incorporates the selector model into LNS by guiding the destroy phase based on predicted node that remain unchanged. Our contributions are summarized as follows:
        
        \begin{enumerate}
            \item A node destroyer selector model that predicts which customer nodes should be removed during the destroy phase of LNS, based on graph-structured input.
            \item An implementation pipeline that enables integration of learned destroy operators into LNS-based search operator.
            \item A hybrid optimization framework that integrates the proposed node-destroyer model with existing metaheuristic baselines to solve the CVRP.
        \end{enumerate}

        The remainder of this chapter is organized as follows: \Cref{sec:problem-definition} outlines the problem addressed in this chapter, while \Cref{sec:baseline-metaheuristic} outlines the baseline metaheuristic algorithms employed in this study and \Cref{sec:proposed-pipeline} details the proposed hybrid pipeline model and \Cref{sec:mechanism} explains its integration strategy to solve the problem. Then, \Cref{sec:experiment} reports the experimental setup and results used to assess our approach, while \Cref{sec:ch5-conclusion} concludes the chapter with final remarks and future directions.

\section{Problem Definition}
\label{sec:problem-definition}
    We focus this study on the Capacitated Vehicle Routing Problem (CVRP), an optimization problem within the broader class of the Vehicle Routing Problems (VRPs). Mathematically, the CVRP can be modeled as an undirected graph $G = (V, E)$, where the set of nodes $V$ includes a depot $c_0$ and a set of customers $C = \{c_1, c_2, \dots, c_N\}$, with $N = |V| - 1$ where every customer has demand $q$. For any pair of nodes $c_i, c_j \in V$, let $d(c_i, c_j)$ denote the distance (or cost) between them. A \textit{route} is defined as a sequence of nodes starting and ending at the depot $c_0$, visiting a subset of customers such that the total demand along the route does not exceed the vehicle capacity $Q$. A solution to the CVRP is a collection of such routes, covering all customers exactly once, and minimizing the total cost (\textit{i.e.}, the sum of distances traveled across all routes) \citep{laporte2009fifty}. A solution is considered \textit{feasible} if every customer node is visited exactly once, and no vehicle exceeds its capacity.

\section{Baseline Metaheuristic}
\label{sec:baseline-metaheuristic}
    Heuristics and local search mechanisms are core components of metaheuristic algorithms \citep{Prodhon2016}. The LNS is a heuristic operator that iteratively improves a solution through destruction and reconstruction phases. Its effectiveness relies on several configurable components, one of which is the selection of \textit{operators}. These include both the number and type of removal operators, which dictate how the solution is disrupted, and insertion operators, which reconstruct the partial solution \citep{VOIGT2025357}. 
    
    Recent developments in LNS operator for the CVRP can be traced back to shared mechanisms found in FILO \citep{accorsi2021fast} and HGS-PILS \citep{arnold2021pils}. In FILO, in particular the adaptive shaking mechanism, where customer nodes are removed and reinserted based on dynamically adjusted disruption levels, improved the ruin-and-recreate mechanism used in SISRs \citep{christiaens2020slack}. Similarly, HGS-PILS \citep{arnold2021pils} incorporates a pattern injection mechanism that guides the search within HGS baseline. Building on this development, the baseline metaheuristic algorithms used in this research are HGS-PILS and FILO.

    \subsection{Hybrid Genetic Search}
    \label{subsec:baseline-1}
        The Hybrid Genetic Search (HGS) \citep{vidal2022hybrid} uses a simple mechanism for generating and utilized a set of local search operator for refining solutions. It begins by creating an initial population of size $\mu$, composed of random solutions that are then enhanced using a set of local search neighborhoods. The initial solutions consist of both feasible and infeasible routes, which are guided by penalties during local search to evolve into better feasible solutions. To strengthen exploration, infeasible solutions are penalized based on their capacity violation rather than discarded.
        
        The \textit{Hybrid Genetic Search with Pattern Injection Local Search} (HGS-PILS) enhances local search neighborhoods in HGS by leveraging frequently visited customer node patterns to guide patter local search in the improvement phase \citep{arnold2021pils}. It operates in two phases: pattern extraction, where recurring customer subsequences from high-quality solutions are identified and stored; and pattern injection, where these patterns are inserted into current solutions by replacing segments and reconnecting remaining routes using a recursive algorithm. This method balances exploiting learned patterns with exploring new configurations, aiming for better local optima without exhaustive search.

    \subsection{Fast Iterated Local Search Localized Optimization}
    \label{subsec:baseline-2}
        The Fast Iterated Local Search Localized Optimization (FILO) metaheuristic solves the large-scale CVRP efficiently \citep{accorsi2021fast}, with FILO2 \citep{ACCORSI2024106562} offering improved scalability. After a construction phase, FILO applies route minimization and a core optimization procedure combining local search and adaptive shaking. The \textit{adaptive shaking mechanism} performs removing and reinserting customer nodes. A shaking parameter $\omega_i$ controls the disruption intensity per customer and is updated based on changing in quality of the resulted solution, in which encouraging more disruption if the outcome is worse, or less disruption if it is almost identical. This structure-aware mechanism targets only relevant regions, balancing exploration and exploitation. 
        Initial $\omega_i$ values depend on instance size and dynamically changes over iterations. The principle of adaptive shaking in FILO closely resembles ruin-and-recreate in SISRs \citep{christiaens2020slack}, as both aim to perturb a solution by removing a subset of customer nodes, balancing exploration and exploitation.

\begin{figure}[htbp]
    \centering
    \includegraphics[width=1\textwidth]{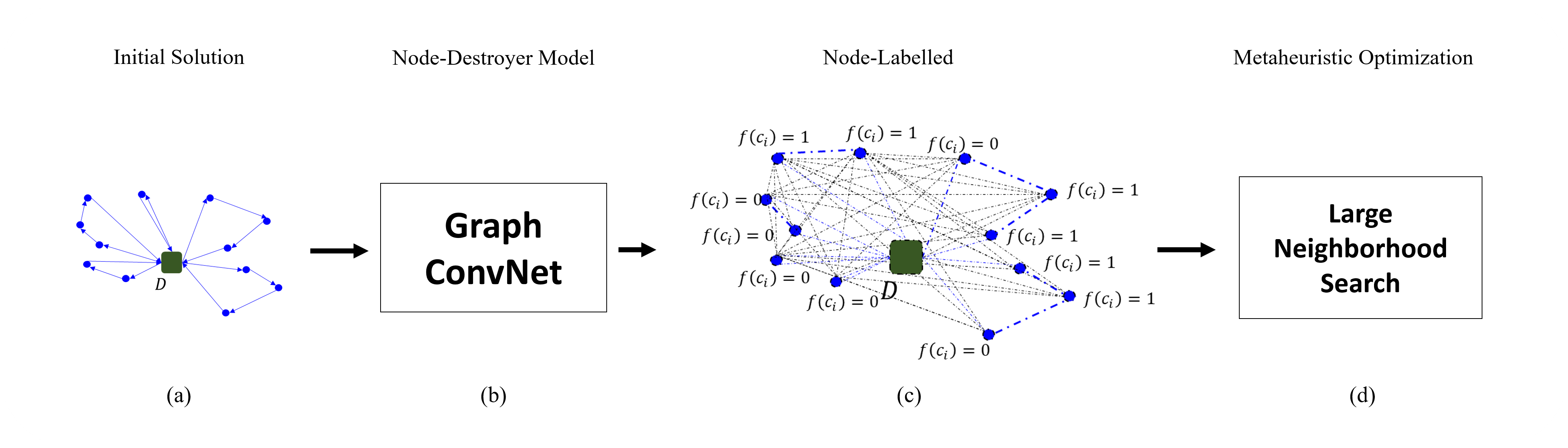}
    \caption{Overview of the proposed hybrid framework.} \label{fig:framework} 
\end{figure}

\section{Proposed Pipeline}
\label{sec:proposed-pipeline}
\begin{figure}[htbp]
    \centering
    \includegraphics[width=1\textwidth]{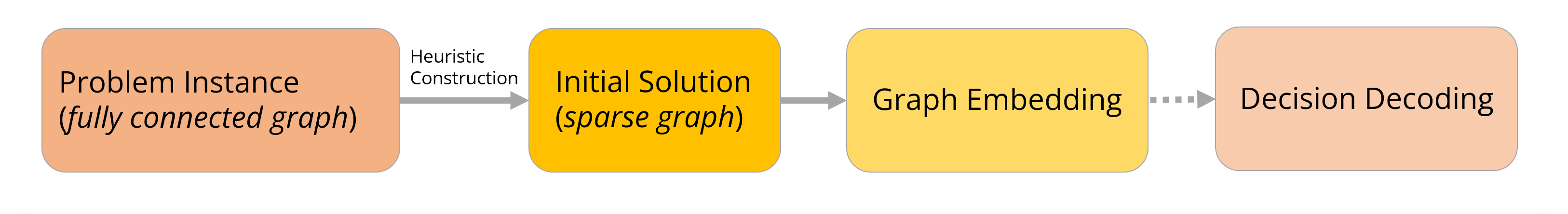}
    \caption{The pipeline of our developed selector $f_{\theta}$.} \label{fig:pipeline} 
\end{figure}

        \Cref{fig:framework} provides an overview of the proposed hybrid optimization framework. The process begins with the construction of an initial solution $S_0$ (a), typically generated using a construction heuristic algorithm. Next, the proposed selector model $f_{\theta}$ (b), which leverages GNNs, assigns binary labels $f_{\theta}(c_i) \in \{0, 1\} \quad \forall i \in 1,\ldots, N$ to all customer nodes in the graph, excluding the depot node $c_0 \notin V(S_0)$. In this research, $f_{\theta}$ is developed using the graph \textit{Convolutional Network} (ConvNet) architecture \citep{joshi2019efficient,kool2022deep}. These labels identify nodes that should be preserved during the LNS procedure. The labeled graph is then passed to the LNS phase (c), where nodes marked by the selector are excluded from the destroy phase, allowing the algorithm to iteratively improve the solution while maintaining several structural components considered \textit{marked} by the model. Additionally, \Cref{fig:pipeline} depicts the overall pipeline proposed in this research. The process starts from a fully connected graph derived from the input problem instance. An initial solution is constructed using a construction heuristic algorithm (depending on the baseline metaheuristic), resulting in a sparse subgraph, as shown in \Cref{fig:framework} (a), in which the initial solution $S_0$. In the next stage, the initial process of the selector $f_{\theta}$ begins by performing embeddings using the graph ConvNet, analogous to step (b) in the framework. 

\begin{figure}[htbp]
    \centering
    \includegraphics[width=0.8\textwidth]{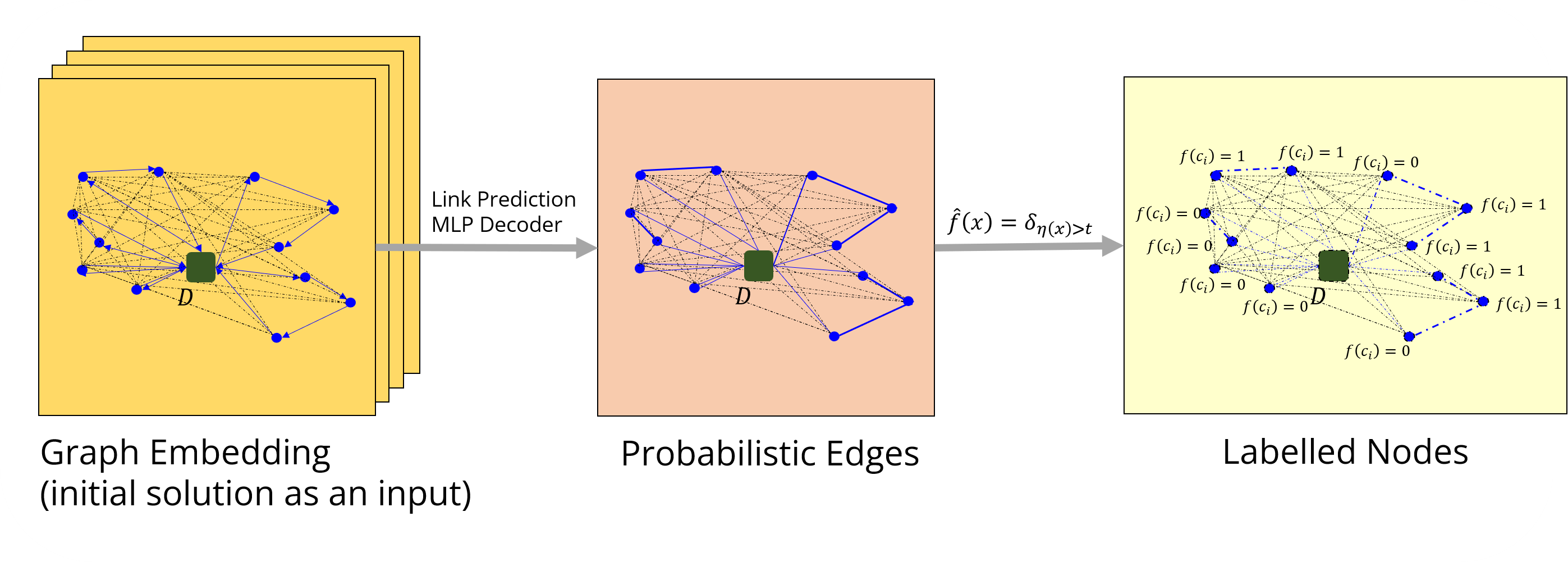}
    \caption{Overview of decision decoding of the selector $f_{\theta}$.} \label{fig:decoding} 
\end{figure}

        This is followed by further process of the selector $f_{\theta}$, in which the decision decoding phase, where each customer node $c_j \in V \setminus \{c_0\}$ receives a binary label $f_{\theta}(c_j) \in \{0,1\}$. A label of $1$ indicates that the node is \textit{marked} and will be prohibited from removal during the destroy phase of the LNS procedure. The decision decoding process is then further detailed in \Cref{fig:decoding}. In contrast to earlier work \citep{joshi2019efficient,kool2022deep,Joshi2022}, which primarily relied on heatmap-based probabilities to predict the likelihood of edges appearing in the optimal solution, our method evaluates the probability of each edge in the initial solution $S_0$ being part of a high-quality final solution. We then perform a binary classification over these edges to determine their relevance. 

        Specifically, we focus on edges in the set $E(S_0) \subseteq E$ and classify whether they should be preserved. If an edge is selected, the corresponding nodes it connects are marked as preserved. These preserved nodes are then exempted from removal during the LNS destroy phase, ensuring that critical parts of the initial solution are preserved while the rest is iteratively improved by the metaheuristic.

    \subsection{Graph Embedding}
    \label{subsec:graph-embedding}
        The graph ConvNet with classical search techniques was initially used to generate approximate TSP solutions directly from problem instances \citep{joshi2019efficient}. In this research, we extend the Graph ConvNet for solving the CVRP. Specifically, we adapt this hybrid mechanism to guide the search process by selecting customer nodes during the removal stage. Each input node feature is represented as a three-dimensional vector $x_i \in [0,1]^3$, encoding the node’s coordinates and its demand utilization $q/Q$. 

\begin{figure}[htbp]
    \begin{center}
        \includegraphics[width=0.42\textwidth]{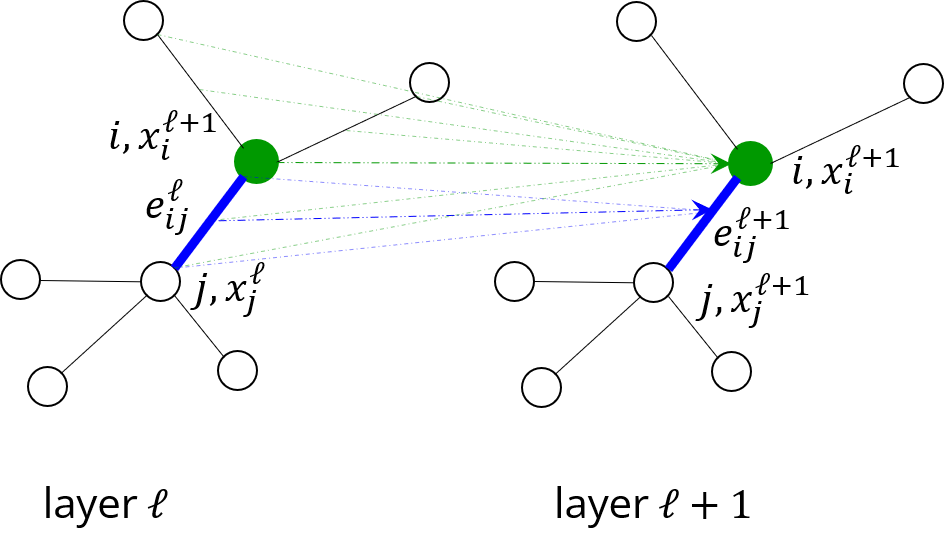}
        \caption{Representation message passing in graph ConvNet.}
        \label{img:node-convnet}
    \end{center}
\end{figure}

        The edge feature, representing the rounded Euclidean distance $d_{ij}$, is embedded as an $h/2$-dimensional vector. Let $x_i^{\ell}$ and $e_{ij}^{\ell}$ denote the node and edge feature vectors at layer $\ell$, respectively. The message-passing update for nodes is defined as:
        \begin{equation}
            x_i^{\ell+1} = x_i^{\ell} + \mathrm{ReLU} \left( \mathrm{BN} \left( W_1^{\ell} x_i^{\ell} + \sum_{j \sim i} \eta_{ij}^{\ell} \odot W_2^{\ell} x_j^{\ell} \right) \right)
        \end{equation}
        with $\eta_{ij}^{\ell} = \frac{\sigma\left( e_{ij}^{\ell} \right)}{\sum_{j' \sim i} \sigma\left( e_{ij'}^{\ell} \right) + \varepsilon}$, where $\varepsilon$ is a small constant to ensure numerical stability. The edge update is then defined as:
        \begin{equation}
            e_{ij}^{\ell+1} = e_{ij}^{\ell} + \mathrm{ReLU} \left( \mathrm{BN} \left( W_3^{\ell} e_{ij}^{\ell} + W_4^{\ell} x_i^{\ell} + W_5^{\ell} x_j^{\ell} \right) \right)
        \end{equation}
        where $W \in \mathbb{R}^{h \times h}$ is a learnable weight matrix. Here, $\sigma$ and $\text{ReLU}$ denote activation functions, while $\text{BN}$ refers to Batch Normalization, applied to normalize activations within each mini-batch. The symbol $\odot$ represents the element-wise (Hadamard) product. The term $\eta_{ij}^{\ell}$ acts as a dense attention map, weighting the contributions of neighboring nodes via a normalized, attention-like score derived from $e_{ij}^{\ell}$. As illustrated in \Cref{img:node-convnet}, the node and edge representations are updated through stacked layers. Green and blue arrows in the figure indicate the flow of information between nodes and edges, enabling structural learning to predict a heatmap of promising edges.
            
    \subsection{Binary Classification}
    \label{subsec:binary-classification}
        The final-layer edge embedding $e_{ij}^L$ is used to predict the probability $p_{ij}^{\text{CVRP}} \in [0, 1]$ of edge $(i, j)$ being part of the CVRP solution, computed via an Multilayer Perceptron (MLP), $p_{ij}^{\text{CVRP}} = \text{MLP}(e_{ij}^{L})$, where $L$ is the number of GNN layers and the MLP consists of $\ell_{\text{mlp}}$ hidden layers. Rather than using the probabilities directly, we apply a binary classification scheme: an edge is considered part of the solution if its predicted probability exceeds a predefined threshold $t$, such as $\hat{f}(x) = \delta_{\eta(x) > t}$.

    \subsection{Supervised Learning Policy}
    \label{subsec:supervised-learning}
\begin{figure}[htbp]
    \centering
    \includegraphics[width=0.86\textwidth]{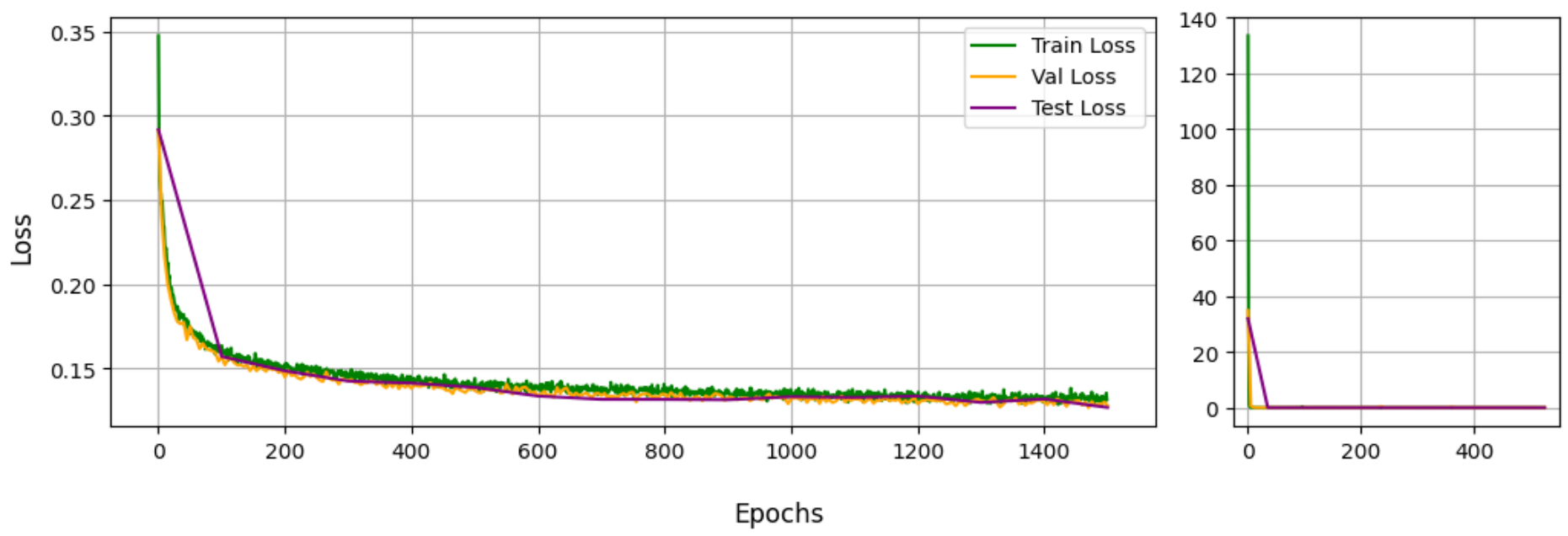}
    \caption{Training loss progression during curriculum learning.} \label{fig:curriculum} 
\end{figure}

        We train our selector model $f_{\theta}$ using CVRP instance-solution pairs of size $N = 100$ from the dataset introduced in \citep{kool2022deep}. These instances are accompanied by supervised labels derived from high-quality solutions generated by the HGS \citep{vidal2022hybrid}, enabling the model to learn meaningful structural patterns. To improve the scalability of $f_{\theta}$ for solving larger and more complex CVRP instances, we adopt a curriculum learning strategy \citep{bengio2009curriculum}, where the model is progressively exposed to more challenging examples during training. Specifically, we fine-tune the selector $f_{\theta}$ on a dataset with $N = 1,000$, generated using the $\mathbb{XML}$100 instance generator\footnote{Available at \url{http://vrp.galgos.inf.puc-rio.br/index.php/en/}} \citep{queiroga202110}. As illustrated in \Cref{fig:curriculum}, the left plot shows the training loss on the initial dataset with $N = 100$, while the right plot shows the fine-tuning loss on the larger dataset with $N = 1,000$. Furthermore, during the training phase, we utilizes the full graph for the CVRP instances with $N = 100$, and a sparse graph for fine-tuning with $k = 25$ nearest neighbors for $N = 1,000$, reducing the size of total edges from $n^2$ to $k \cdot n$.

    \subsection{Graph sparsification}
    \label{subsec:graph-sparsification}
        While the training phase utilizes the full graph for the CVRP instances with $N = 100$, and a sparse graph with $k = 25$ nearest neighbors for $N = 1,000$, the inference phase adopts a different approach. Specifically, during inference, graph evaluation is restricted to those edge forming the initial solution $E(S_0) \subseteq E$, generated by the baseline metaheuristic. In other words, the model operates on a sparse graph, derived from $S_0$. Then, when an edge is selected, its connected nodes are marked as preserved. These nodes are excluded from the removal process during the LNS destroy phase, allowing key components of the initial solution to remain intact while the rest is further optimized.

\section{Hybrid Mechanism}
\label{sec:mechanism}
    As briefly mentioned earlier, we leverage the predicted probability of each edge being part of the final CVRP solution as the basis for a binary classification task, as described in \Cref{subsec:binary-classification}. In this process, if an edge is classified as part of the hight-quality solution (\textit{i.e.}, $\hat{f}(x) = 1$), the nodes it connects are preserved to the current solution structure. Consequently, this pair of nodes is marked as prohibited and excluded from selection during the destroy phase of the LNS procedure. This mechanism ensures that certain strategically important nodes are retained throughout the optimization process. By preserving these nodes, the LNS can focus its search on the remaining parts of the solution space
    
    \subsection{Hybrid Mechanism for HGS-PILS}
    \label{subsec:hybrid-mechanism-1}
        In HGS-PILS, the hybrid mechanism leverages node-level information from the selector $f_{\theta}$ to exclude specific nodes during pattern-based operations. Nodes marked by the selector $f_{\theta}$ are considered protected and are not altered throughout the optimization process. During each iteration, every node within a candidate pattern is compared against this list of marked nodes. If a match is detected, that node is skipped during key steps such as marking, route selection, and segment construction. This selective filtering ensures that critical components of the current solution remain preserved, while the remaining nodes can be modified.
        \begin{enumerate}
            \item Use the selector $f_{\theta}$ to mark specific nodes as \textit{prohibited}.
            \item During pattern-based operations, check each node in the candidate pattern against the list of marked nodes.
            \item If a node is marked, skip it during marking, route selection, and segment building.
            \item Ensure that only unmarked nodes are involved in pattern operations, preserving the fixed status of marked nodes.
            \item Apply the modified pattern to the rest of the solution without altering marked nodes.
        \end{enumerate} 
    
    \subsection{Hybrid Mechanism for FILO2}
    \label{subsec:hybrid-mechanism-2}
        In FILO2, the hybrid mechanism is implemented through a filtering procedure that excludes specific prohibited nodes, those identified and marked by the selector $f_{\theta}$, during the ruin phase. This selective exclusion ensures that the number of customer nodes removed remains consistent, and the criteria for selecting the next node to ruin are preserved. As a result, the marked nodes are kept fixed throughout the process, while the remaining nodes are eligible for modification, allowing the optimization to proceed without disrupting key structural components of the initial solution.
        \begin{enumerate}
            \item Use the selector $f_{\theta}$ to identify and mark specific nodes as \textit{prohibited}.
            \item During the ruin process, apply a filtering procedure to exclude these prohibited nodes.
            \item Ensure that the total number of customer nodes removed remains unchanged.
            \item Maintain the existing logic for selecting the next node to ruin.
            \item Allow only unmarked nodes to be modified, while keeping marked nodes fixed.
        \end{enumerate}

        \paragraph{\textbf{Solving very large scale problems}} Sparse graphs can accelerate the learning and inference process without significantly compromising prediction quality \citep{kool2022deep}. For very large instances (\textit{e.g.}, $N > 1000$), using a full graph becomes computationally expensive due to the exponential increase in number of edges. To address this, model inference focuses only on the edges forming the initial solution $S_0$. Thus, for very large-scale of problems, only the $1,000$ customer nodes nearest to the depot $c_0$ are considered when performing the selector $f_{\theta}$. This approach has been tested during model development through empirical evaluation, demonstrating its ability to reduce computational load while preserving prediction accuracy.

\section{Computational Experiment}
\label{sec:experiment}
    In this section, we report the computational experiment results of integrating the proposed hybrid mechanism into baseline metaheuristics. The selector model $f_{\theta}$ was implemented in Python\footnote{Available at \url{https://github.com/bachtiarherdianto/MH-Node-Destroyer}}, while the baseline metaheuristics were implemented in C++. The sourcecode for both HGS-PILS\footnote{Available at \url{https://w1.cirrelt.ca/~vidalt/en/VRP-resources.html}} and FILO2\footnote{Available at \url{https://github.com/acco93/filo2}} is publicly available. We modified the original sourcode of both HGS-PILS and FILO2 to incorporate the proposed hybrid iterative learning guidance, which adds ML guidance within the iterative loop of the algorithms.

    \paragraph{\textbf{Training the selector}}
        The training of the baseline graph model was conducted on a 64-bit Debian GNU/Linux 12 machine with 16 virtual AMD cores, 62 GB RAM, and an NVIDIA L40 GPU. During evaluation on the validation and test sets, the adjacency matrix representing the probability of all edges in the initial solution which obtained via the graph ConvNet is then converted into binary decisions. Using the high quality solutions as ground-truth labels, we compute the precision metric as described in \Cref{subsec:binary-classification} and illustrated in \Cref{fig:decoding}. The architecture of $f_{\theta}$ consist of $\ell_{conv}=10$ graph layer and $\ell_{mlp}=3$ layers in the $\text{MLP}$ with hidden dimension $h=120$ for each layer. The selector model $f_{\theta}$ was initially trained using an NVIDIA L40 GPU on CVRP instances with $N = 100$, leveraging high-quality labels derived from HGS solutions \citep{kool2022deep}. This training phase lasted approximately $1.6$ hours and used $2$ GB of GPU memory. To scale $f_{\theta}$ to more complex problems, we adopted a curriculum learning approach by fine-tuning the model on larger instances with $N = 1,000$, as described in the previous section. This second phase consumed $42.9$ GB of GPU memory and took approximately $7.5$ hours. The curriculum strategy enhanced the robustness of the prediction of the model and improved its effectiveness on large-scale instances.

    \paragraph{\textbf{Hyperparameter setup}}
        To evaluate the performance of the hybrid optimization, we adopted experimental settings from previous studies. For HGS-PILS, we followed the original HGS configuration \citep{vidal2022hybrid}, setting the maximum runtime as $T_{\text{max}} = N \times 2.4$ seconds for $\mathbb{X}$ instances. For the hybrid FILO2, we used the original FILO2 configuration \citep{ACCORSI2024106562}, where the maximum number of core optimization iterations is $\Delta_{\text{CO}} = 10^5$. The binary classification threshold $t$ for decision decoding was set to 0.8 for HGS-PILS. For FILO2, $t = 0.9$ was used for instances with $N < 300$, and $t = 0.85$ for $N \geq 300$. Additionally, we incorporated a tabu aspiration mechanism \citep{glover1997tabu} using a randomized threshold parameter $p_\Theta$. Marked nodes were permitted to move if the randomly sampled value $\mathcal{R} \sim \mathcal{U}(0, 1)$ satisfied $\mathcal{R} > p_\Theta$. The values of $p_\Theta$ were determined based on empirical analysis through model development. Specifically, for HGS-PILS solving $\mathbb{X}$ instances, $p_\Theta$ was set to $0.65$ for $N < 300$ and $0.7$ for larger instances. For FILO2 solving $\mathbb{X}$ instances, we set $p_\Theta$ to $0.6$ for smaller cases and $0.65$ for larger ones. When solving $\mathbb{B}$ instances, a fixed $p_\Theta = 0.6$ was used for FILO2.

    \paragraph{\textbf{Experimental setup}}
        The optimization experiments were also carried out on the same 64-bit Debian GNU/Linux 12 machine with 16 virtual AMD cores, 62 GB RAM, and an NVIDIA L40 GPU. Due to the stochastic nature of the algorithm that used pseudo-random generator \citep{matsumoto1998mersenne}, each experiment was repeated five times. The random seed for each run was set as the run index minus one. Throughout the experimentation, we refer to the following:
    
        \begin{itemize}
            \item \textbf{BKS:} The total cost value of the best-known solution. All instance details are available at \url{http://vrp.galgos.inf.puc-rio.br/index.php/en/}.
            \item \textbf{Gap:} The relative difference between the obtained solution and the optimal (or best-known) solution / BKS, calculated as:
            \begin{equation}
                \text{Gap} = \dfrac{\text{Obtained Solution} - \text{BKS}}{\text{BKS}} \times 100\%
            \end{equation}
        \end{itemize}

    \paragraph{\textbf{Statistical Analysis}}
        To evaluate the effectiveness of the proposed mechanism, we performed a non-parametric one-tailed Wilcoxon signed-rank test \citep{JMLR:v7:demsar06a,arnold2021pils,accorsi2022guidelines,zarate2025machine}. Throughout the experimentation, the following hypotheses were formulated:
    
        \setcounter{hyp}{-1}
        \begin{hyp} \label{hyp:ch5-a}
            There is no significant difference between the baseline and the hybrid mechanism.
        \end{hyp}
        \begin{hyp} \label{hyp:ch5-b}
            There is a significant difference between the baseline and the hybrid mechanism.
        \end{hyp}
    
        Throughout the experimentation, we set the significance level to $\alpha = 0.05$ \citep{arnold2021pils,zarate2025machine}. If the resulting $p$-value is less than or equal to $\alpha$, we reject \Cref{hyp:ch5-a}, thereby concluding that the observed performance differences are statistically significant and that the proposed hybrid mechanism performs significantly better than the baseline.

\begin{table}[htbp] 
    \begin{center}
        \caption{Effect of hybrid node-destroyer on $\mathbb{X}$ instances.}
        \label{table:summary-x-result}
        \vspace*{0.2cm}
        \setlength\tabcolsep{3pt}
        \scalebox{0.66}{
            \begin{tabular}{l rr rr rrr rrr}
                \toprule
                \multirow{2}{*}{\textbf{Problem Size}} 
                    & \multicolumn{2}{c}{\textbf{HGS-PILS}} 
                    & \multicolumn{2}{c}{\textbf{HGS-PILS$^{*}$}} 
                    & \multicolumn{3}{c}{\textbf{FILO2}} 
                    & \multicolumn{3}{c}{\textbf{FILO2$^{*}$}} \\
                \cmidrule(lr){2-3} \cmidrule(lr){4-5} \cmidrule(lr){6-8} \cmidrule(lr){9-11}
                & \textbf{Avg Gap} & \textbf{Best Gap} 
                & \textbf{Avg Gap} & \textbf{Best Gap} 
                & \textbf{Avg Gap} & \textbf{Best Gap} & \textbf{Time (s)} 
                & \textbf{Avg Gap} & \textbf{Best Gap} & \textbf{Time (s)} \\
                \midrule
                100 -- 200     & \textbf{0.0294} & 0.0104 & 0.0306 & 0.0081 & 0.1384 & 0.0474 & 91.26 & \textbf{0.1338} & 0.0571 & 99.00 \\
                204 -- 491     & 0.2457          & 0.1400 & \textbf{0.2358} & 0.1294 & \textbf{0.3105} & 0.1889 & 88.82 & 0.3415 & 0.2975 & 105.98\\
                502 -- 749     & 0.4982          & 0.3786 & \textbf{0.4929} & 0.3490 & \textbf{0.4630} & 0.3579 & 87.37 & 0.4674 & 0.4326 & 104.56 \\
                766 -- 1001    & 0.6701          & 0.5469 & \textbf{0.6567} & 0.5351 & \textbf{0.5248} & 0.3886 & 22.4 & 0.5436 & 0.4907 & 104.60 \\
                \midrule
                Average Gap    & 0.3012          &  & \textbf{0.2942} &  & \textbf{0.3311} &  & 89.02 & 0.3460 &  & 104.00 \\
                \bottomrule
            \end{tabular}
        }
    \end{center}
\end{table}

    \paragraph{\textbf{Computational results on large instances}}
        We first evaluate the proposed hybrid mechanism on the $\mathbb{X}$ instances \citep{UCHOA2017845}. A comparative analysis is conducted between the baseline algorithms and their hybrid versions, denoted by a superscript $^*$. The results are summarized in \Cref{table:generalized-hgspils-detailed-1}, which presents the performance across different problem sizes. In general, the hybrid mechanism either maintains or slightly improves solution quality. Specifically, HGS-PILS$^{*}$, as described in \cref{table:summary-x-result}, shows marginal improvements in the best gap values across most instance sizes compared to the original HGS-PILS. In contrast, the hybrid variant of FILO2 does not consistently outperform the baseline. Noticeable gains are only observed for smaller instances with fewer than $200$ customer nodes. On average, across the $\mathbb{X}$ instances, the hybrid mechanism performs competitively better, particularly for the HGS-PILS$^{*}$.

\begin{table}[htbp] 
    \begin{center}
        \caption{Effect of hybrid node-destroyer on $\mathbb{X}$ instances based on depot positions.}
        \label{table:summary-x-result-depot-pos}
        \vspace*{0.2cm}
        \setlength\tabcolsep{4pt}
        \scalebox{0.8}{
            \begin{tabular}{ll l l c l l}
                \toprule
                \textbf{Depot Position} 
                && \textbf{HGS-PILS} 
                & \textbf{HGS-PILS$^{*}$} 
                && \textbf{FILO2} 
                & \textbf{FILO2$^{*}$} \\
                \midrule
                Central (C) && 0.291 & \textbf{0.281} && 0.322 & \textbf{0.307} \\
                Edge (E) && \textbf{0.297} & 0.299 && \textbf{0.333} & 0.366 \\
                Random (R) && 0.310 & \textbf{0.300} && \textbf{0.336} & 0.343 \\
                \bottomrule
            \end{tabular}
        }
    \end{center}
\end{table}

    \Cref{table:summary-x-result-depot-pos} further breaks down the results according to depot positioning. The depot positions in the $\mathbb{X}$ instances are categorized as Central (C), Edge (E), and Random (R) \citep{UCHOA2017845}. For HGS-PILS and HGS-PILS$^{*}$, the results remain relatively stable across all depot configurations. Conversely, FILO2 exhibits greater variability. Its hybrid version, FILO2$^{*}$, shows a slight improvement in average gap for centrally located depots, indicating that the node-destroyer model is more effective in such configurations. 

\begin{figure}[htbp]
    \centering
    \includegraphics[width=1\textwidth]{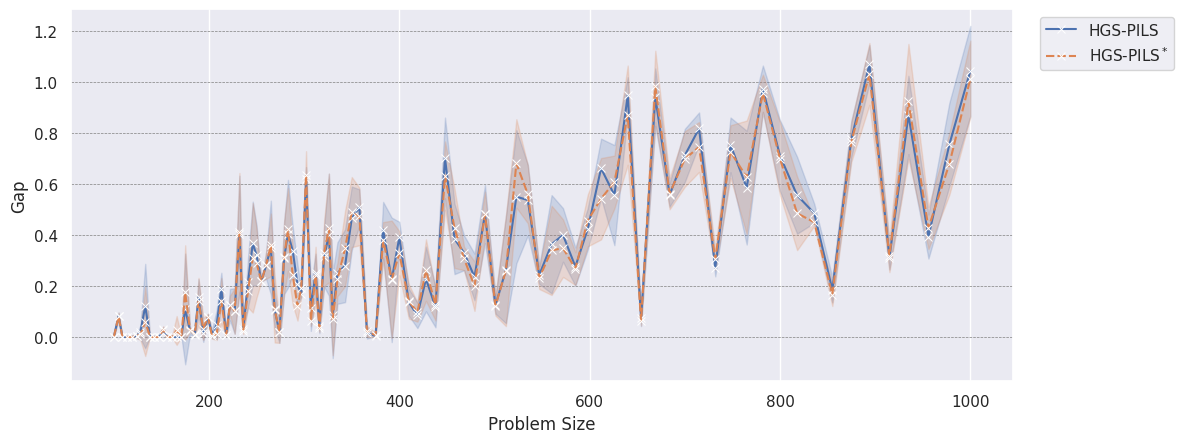}
    \caption{Comparison performance of HGS-PILS on $\mathbb{X}$ instances.} \label{fig:hgs-pils-x} 
\end{figure}

\begin{figure}[htbp]
    \centering
    \includegraphics[width=1\textwidth]{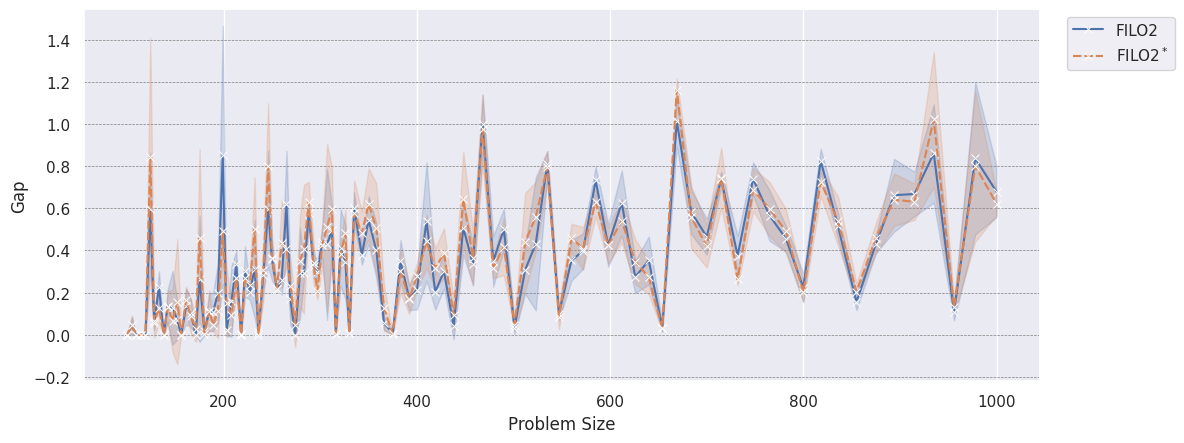}
    \caption{Comparison performance of FILO2 on $\mathbb{X}$ instances.} \label{fig:filo2-x} 
\end{figure}

    However, for the Edge and Random depot scenarios, the hybrid mechanism does not outperform the baseline. The detailed performance trends are illustrated in \Cref{fig:hgs-pils-x} for HGS-PILS and \Cref{fig:filo2-x} for FILO2. Finally, shown in \Cref{fig:hgs-pils-x} for baseline HGS-PILS, on $\mathbb{X}$ instances, the one-tailed Wilcoxon signed-rank test rejects \Cref{hyp:ch5-a} with $p$-value $= 0.033$ $(< 0.05)$, indicating that proposed hybrid node destroyer statistically improve the performance of the baseline HGS-PILS when solving $\mathbb{X}$ instances.

\begin{table}[htbp]
    \begin{center}
        \caption{Detailed performances HGS-PILS on $\mathbb{X}$ instances.}
        \label{table:generalized-hgspils-detailed-1}
        \vspace*{0.2cm}
        \setlength\tabcolsep{4pt}
        \scalebox{0.7}{
            \begin{tabular}{ll rr r rr rr}
                \toprule
                    \multirow{2}{*}{\textbf{Instance}} 
                    && \multicolumn{2}{c}{\textbf{HGS-PILS}} 
                    && \multicolumn{2}{c}{\textbf{HGS-PILS$^{*}$}}
                    && \multirow{2}{*}{\textbf{BKS}}\\
                    \cmidrule(lr){3-4} \cmidrule(lr){6-7}
                    && \textbf{Avg Gap} & \textbf{Best Gap} 
                    && \textbf{Avg Gap} & \textbf{Best Gap} \\
                \midrule
                    X-n101-k25 && \textbf{27591 (0.000)} & 27591 (0.000) && \textbf{27591 (0.000)} & 27591 (0.000) && 27591\\
                    X-n106-k14 && 26383.8 (0.083) & 26378 (0.061) && \textbf{26383.4 (0.081)} & 26378 (0.061) && 26362\\
                    X-n110-k13 && \textbf{14971 (0.000)} & 14971 (0.000) && \textbf{14971 (0.000)} & 14971 (0.000) && 14971\\
                    X-n115-k10 && \textbf{12747 (0.000)} & 12747 (0.000) && \textbf{12747 (0.000)} & 12747 (0.000) && 12747\\
                    X-n120-k6 && \textbf{13332 (0.000)} & 13332 (0.000) && \textbf{13332 (0.000)} & 13332 (0.000) && 13332\\
                    X-n125-k30 && \textbf{55541 (0.004)} & 55539 (0.000) && \textbf{55541 (0.004)} & 55539 (0.000) && 55539\\
                    X-n129-k18 && 28945.6 (0.019) & 28940 (0.000) && \textbf{28942.8 (0.010)} & 28940 (0.000) && 28940\\
                    X-n134-k13 && 10929.2 (0.121) & 10916 (0.000) && \textbf{10922.6 (0.060)} & 10916 (0.000) && 10916\\
                    X-n139-k10 && \textbf{13590 (0.000)} & 13590 (0.000) && \textbf{13590 (0.000)} & 13590 (0.000) && 13590\\
                    X-n143-k7 && \textbf{15700 (0.000)} & 15700 (0.000) && \textbf{15700 (0.000)} & 15700 (0.000) && 15700\\
                    X-n148-k46 && \textbf{43448 (0.000)} & 43448 (0.000) && \textbf{43448 (0.000)} & 43448 (0.000) && 43448\\
                    X-n153-k22 && \textbf{21226 (0.028)} & 21225 (0.024) && 21226.2 (0.029) & 21225 (0.024) && 21220\\
                    X-n157-k13 && \textbf{16876 (0.000)} & 16876 (0.000) && \textbf{16876 (0.000)} & 16876 (0.000) && 16876\\
                    X-n162-k11 && \textbf{14138 (0.000)} & 14138 (0.000) && \textbf{14138 (0.000)} & 14138 (0.000) && 14138\\
                    X-n167-k10 && \textbf{20557 (0.000)} & 20557 (0.000) && 20562.2 (0.025) & 20557 (0.000) && 20557\\
                    X-n172-k51 && \textbf{45607 (0.000)} & 45607 (0.000) && \textbf{45607 (0.000)} & 45607 (0.000) && 45607\\
                    X-n176-k26 && \textbf{47864.4 (0.110)} & 47812 (0.000) && 47895.4 (0.174) & 47812 (0.000) && 47812\\
                    X-n181-k23 && \textbf{25575.2 (0.024)} & 25569 (0.000) && 25576.2 (0.028) & 25573 (0.016) && 25569\\
                    X-n186-k15 && \textbf{24147 (0.008)} & 24145 (0.000) && 24147.8 (0.012) & 24145 (0.000) && 24145\\
                    X-n190-k8 && 17006.4 (0.155) & 16994 (0.082) && \textbf{17003.6 (0.139)} & 16983 (0.018) && 16980\\
                    X-n195-k51 && \textbf{44232 (0.016)} & 44225 (0.000) && 44239 (0.032) & 44225 (0.000) && 44225\\
                    X-n200-k36 && \textbf{58624 (0.079)} & 58614 (0.061) && \textbf{58624 (0.079)} & 58614 (0.061) && 58578\\
                    X-n204-k19 && 19567 (0.010) & 19565 (0.000) && \textbf{19566 (0.005)} & 19565 (0.000) && 19565\\
                    X-n209-k16 && 30670.6 (0.048) & 30656 (0.000) && \textbf{30659 (0.010)} & 30656 (0.000) && 30656\\
                    X-n214-k11 && 10876.6 (0.190) & 10867 (0.101) && \textbf{10871.4 (0.142)} & 10858 (0.018) && 10856\\
                    X-n219-k73 && \textbf{117601.6 (0.006)} & 117595 (0.000) && \textbf{117601.6 (0.006)} & 117595 (0.000) && 117595\\
                    X-n223-k34 && 40488.2 (0.127) & 40463 (0.064) && \textbf{40487.4 (0.125)} & 40463 (0.064) && 40437\\
                    X-n228-k23 && \textbf{25768.2 (0.102)} & 25743 (0.004) && \textbf{25768.2 (0.102)} & 25743 (0.004) && 25742\\
                    X-n233-k16 && \textbf{19307.6 (0.404)} & 19230 (0.000) && 19309.2 (0.412) & 19230 (0.000) && 19230\\
                    X-n237-k14 && 27050.6 (0.032) & 27050 (0.030) && \textbf{27048.8 (0.025)} & 27042 (0.000) && 27042\\
                    X-n242-k48 && 82927.8 (0.214) & 82867 (0.140) && \textbf{82900.6 (0.181)} & 82867 (0.140) && 82751\\
                    X-n247-k50 && 37410.8 (0.367) & 37309 (0.094) && \textbf{37389.6 (0.310)} & 37298 (0.064) && 37274\\
                    X-n251-k28 && 38807.6 (0.320) & 38732 (0.124) && \textbf{38796 (0.290)} & 38716 (0.083) && 38684\\
                    X-n256-k16 && \textbf{18880 (0.218)} & 18880 (0.218) && \textbf{18880 (0.218)} & 18880 (0.218) && 18839\\
                    X-n261-k13 && \textbf{26632.4 (0.280)} & 26612 (0.203) && \textbf{26632.4 (0.280)} & 26612 (0.203) && 26558\\
                    X-n266-k58 && \textbf{75742.6 (0.351)} & 75608 (0.172) && 75750.2 (0.361) & 75635 (0.208) && 75478\\
                    X-n270-k35 && \textbf{35328.8 (0.107)} & 35309 (0.051) && 35329.2 (0.108) & 35303 (0.034) && 35291\\
                    X-n275-k28 && \textbf{21249 (0.019)} & 21245 (0.000) && \textbf{21249 (0.019)} & 21245 (0.000) && 21245\\
                    X-n280-k17 && \textbf{33607 (0.310)} & 33524 (0.063) && 33608.2 (0.314) & 33524 (0.063) && 33503\\
                    X-n284-k15 && \textbf{20296.8 (0.405)} & 20260 (0.223) && 20300.2 (0.421) & 20275 (0.297) && 20215\\
                    X-n289-k60 && 95472.2 (0.338) & 95352 (0.211) && \textbf{95381.6 (0.242)} & 95228 (0.081) && 95151\\
                    X-n294-k50 && 47258 (0.206) & 47222 (0.129) && \textbf{47217.8 (0.120)} & 47171 (0.021) && 47161\\
                    X-n298-k31 && \textbf{34288.2 (0.167)} & 34284 (0.155) && 34296.4 (0.191) & 34284 (0.155) && 34231\\
                    X-n303-k21 && \textbf{21870.6 (0.619)} & 21855 (0.547) && 21874.4 (0.637) & 21842 (0.488) && 21736\\
                    X-n308-k13 && 25888.2 (0.113) & 25866 (0.027) && \textbf{25875.2 (0.063)} & 25870 (0.043) && 25859\\
                    X-n313-k71 && 94278.6 (0.251) & 94214 (0.182) && \textbf{94270.6 (0.242)} & 94152 (0.116) && 94043\\
                    X-n317-k53 && 78390 (0.045) & 78363 (0.010) && \textbf{78381 (0.033)} & 78363 (0.010) && 78355\\
                    X-n322-k28 && \textbf{29927.8 (0.314)} & 29868 (0.114) && 29932 (0.328) & 29868 (0.114) && 29834\\
                    X-n327-k20 && \textbf{27642.4 (0.401)} & 27577 (0.163) && 27649.6 (0.427) & 27577 (0.163) && 27532\\
                    X-n331-k15 && 31128.4 (0.085) & 31105 (0.010) && \textbf{31123.6 (0.069)} & 31103 (0.003) && 31102\\
                    X-n336-k84 && 139459.6 (0.251) & 139297 (0.134) && \textbf{139424.4 (0.225)} & 139297 (0.134) && 139111\\
                    X-n344-k43 && \textbf{42167.4 (0.279)} & 42110 (0.143) && 42196 (0.347) & 42124 (0.176) && 42050\\
                \bottomrule   
            \end{tabular}
        }
    \end{center}
\end{table}

\begin{table}[htbp]
    \begin{center}
        \caption{Detailed performances HGS-PILS on $\mathbb{X}$ instances (continue).}
        \label{table:generalized-hgspils-detailed-2}
        \vspace*{0.2cm}
        \setlength\tabcolsep{4pt}
        \scalebox{0.7}{
            \begin{tabular}{ll rr r rr rr}
                \toprule
                    \multirow{2}{*}{\textbf{Instance}} 
                    && \multicolumn{2}{c}{\textbf{HGS-PILS}} 
                    && \multicolumn{2}{c}{\textbf{HGS-PILS$^{*}$}}
                    && \multirow{2}{*}{\textbf{BKS}}\\
                    \cmidrule(lr){3-4} \cmidrule(lr){6-7}
                    && \textbf{Avg Gap} & \textbf{Best Gap} 
                    && \textbf{Avg Gap} & \textbf{Best Gap} \\
                \midrule
                    X-n351-k40 && \textbf{26016.4 (0.465)} & 25981 (0.328) && 26022.8 (0.490) & 25985 (0.344) && 25896\\
                    X-n359-k29 && 51766.8 (0.508) & 51716 (0.410) && \textbf{51748.2 (0.472)} & 51685 (0.349) && 51505\\
                    X-n367-k17 && \textbf{22817.6 (0.016)} & 22814 (0.000) && 22819 (0.022) & 22814 (0.000) && 22814\\
                    X-n376-k94 && \textbf{147718 (0.003)} & 147718 (0.003) && 147723.4 (0.007) & 147713 (0.000) && 147713\\
                    X-n384-k52 && 66217.4 (0.421) & 66140 (0.303) && \textbf{66189.4 (0.378)} & 66140 (0.303) && 65940\\
                    X-n393-k38 && \textbf{38345.6 (0.224)} & 38277 (0.044) && 38346.4 (0.226) & 38277 (0.044) && 38260\\
                    X-n401-k29 && 66421.6 (0.391) & 66383 (0.333) && \textbf{66380.2 (0.328)} & 66292 (0.195) && 66163\\
                    X-n411-k19 && \textbf{19738.8 (0.136)} & 19726 (0.071) && 19739.4 (0.139) & 19725 (0.066) && 19712\\
                    X-n420-k130 && \textbf{107889.2 (0.085)} & 107831 (0.031) && 107903.2 (0.098) & 107842 (0.041) && 107798\\
                    X-n429-k61 && \textbf{65597.8 (0.227)} & 65530 (0.124) && 65620.6 (0.262) & 65530 (0.124) && 65449\\
                    X-n439-k37 && \textbf{36433.6 (0.117)} & 36402 (0.030) && 36437 (0.126) & 36416 (0.069) && 36391\\
                    X-n449-k29 && 55620.6 (0.702) & 55490 (0.465) && \textbf{55582.4 (0.633)} & 55503 (0.489) && 55233\\
                    X-n459-k26 && \textbf{24231.8 (0.384)} & 24203 (0.265) && 24241.8 (0.426) & 24213 (0.307) && 24139\\
                    X-n469-k138 && 222555.6 (0.330) & 222392 (0.256) && \textbf{222506.4 (0.308)} & 222386 (0.253) && 221824\\
                    X-n480-k70 && 89659.4 (0.235) & 89530 (0.091) && \textbf{89627 (0.199)} & 89522 (0.082) && 89449\\
                    X-n491-k59 && \textbf{66807.4 (0.482)} & 66736 (0.375) && \textbf{66807.4 (0.482)} & 66745 (0.388) && 66487\\
                    X-n502-k39 && 69311.8 (0.124) & 69291 (0.094) && \textbf{69310.6 (0.122)} & 69282 (0.081) && 69226\\
                    X-n513-k21 && 24264.6 (0.263) & 24201 (0.000) && \textbf{24264 (0.260)} & 24201 (0.000) && 24201\\
                    X-n524-k153 && \textbf{155443 (0.550)} & 154911 (0.206) && 155644.8 (0.680) & 155189 (0.386) && 154593\\
                    X-n536-k96 && \textbf{95377.2 (0.537)} & 95240 (0.392) && 95399.4 (0.560) & 95262 (0.415) && 94868\\
                    X-n548-k50 && 86910.6 (0.243) & 86869 (0.195) && \textbf{86900 (0.231)} & 86843 (0.165) && 86700\\
                    X-n561-k42 && \textbf{42872.2 (0.363)} & 42781 (0.150) && 42862 (0.339) & 42772 (0.129) && 42717\\
                    X-n573-k30 && 50875.6 (0.400) & 50837 (0.324) && \textbf{50849.4 (0.348)} & 50796 (0.243) && 50673\\
                    X-n586-k159 && 190840.6 (0.276) & 190687 (0.195) && \textbf{190825.6 (0.268)} & 190674 (0.188) && 190316\\
                    X-n599-k92 && \textbf{108910.2 (0.423)} & 108833 (0.352) && 108946 (0.456) & 108797 (0.319) && 108451\\
                    X-n613-k62 && 59928.6 (0.661) & 59861 (0.548) && \textbf{59857.6 (0.542)} & 59703 (0.282) && 59535\\
                    X-n627-k43 && \textbf{62509.6 (0.556)} & 62339 (0.282) && 62541.6 (0.607) & 62452 (0.463) && 62164\\
                    X-n641-k35 && 64297.4 (0.947) & 64242 (0.860) && \textbf{64247.8 (0.869)} & 64046 (0.553) && 63694\\
                    X-n655-k131 && 106857.4 (0.072) & 106817 (0.035) && \textbf{106846.8 (0.063)} & 106828 (0.045) && 106780\\
                    X-n670-k130 && \textbf{147708 (0.940)} & 147601 (0.867) && 147769 (0.982) & 147603 (0.869) && 146332\\
                    X-n685-k75 && \textbf{68585 (0.557)} & 68553 (0.510) && 68588.4 (0.562) & 68539 (0.490) && 68205\\
                    X-n701-k44 && 82506.2 (0.712) & 82380 (0.558) && \textbf{82494.6 (0.698)} & 82347 (0.518) && 81923\\
                    X-n716-k35 && 43741.6 (0.817) & 43711 (0.747) && \textbf{43710.2 (0.745)} & 43654 (0.615) && 43387\\
                    X-n733-k159 && \textbf{136558.2 (0.270)} & 136513 (0.237) && 136605.2 (0.305) & 136526 (0.247) && 136190\\
                    X-n749-k98 && 77896.2 (0.753) & 77811 (0.643) && \textbf{77875.4 (0.726)} & 77797 (0.625) && 77314\\
                    X-n766-k71 && \textbf{115124 (0.585)} & 114930 (0.416) && 115170.6 (0.626) & 114928 (0.414) && 114454\\
                    X-n783-k48 && 73098.2 (0.973) & 73002 (0.840) && \textbf{73086.2 (0.956)} & 73000 (0.837) && 72394\\
                    X-n801-k40 && 73825.8 (0.710) & 73731 (0.581) && \textbf{73817.4 (0.699)} & 73740 (0.593) && 73305\\
                    X-n819-k171 && 158999.2 (0.555) & 158785 (0.420) && \textbf{158887.4 (0.485)} & 158637 (0.326) && 158121\\
                    X-n837-k142 && 194676.2 (0.485) & 194576 (0.433) && \textbf{194608.6 (0.450)} & 194576 (0.433) && 193737\\
                    X-n856-k95 && 89129.4 (0.185) & 89071 (0.119) && \textbf{89112.6 (0.166)} & 89071 (0.119) && 88965\\
                    X-n876-k59 && 100069.4 (0.776) & 100011 (0.717) && \textbf{100059.4 (0.766)} & 99994 (0.700) && 99299\\
                    X-n895-k37 && 54436 (1.069) & 54379 (0.964) && \textbf{54415.4 (1.031)} & 54367 (0.941) && 53860\\
                    X-n916-k207 && \textbf{330202.6 (0.311)} & 329965 (0.239) && 330221 (0.317) & 329952 (0.235) && 329179\\
                    X-n936-k151 && \textbf{133880.2 (0.870)} & 133599 (0.659) && 133950.6 (0.923) & 133601 (0.660) && 132725\\
                    X-n957-k87 && \textbf{85801.4 (0.394)} & 85723 (0.302) && 85830.2 (0.427) & 85747 (0.330) && 85465\\
                    X-n979-k58 && 119885.2 (0.755) & 119630 (0.540) && \textbf{119794.8 (0.679)} & 119582 (0.500) && 118987\\
                    X-n1001-k43 && 73113.8 (1.043) & 72996 (0.880) && \textbf{73091 (1.012)} & 72986 (0.867) && 72359\\
                \midrule
                    Average Gap && 0.3012 & && \textbf{0.2942} & && \\
                \bottomrule    
            \end{tabular}
        }
    \end{center}
\end{table}

    \paragraph{\textbf{Computational results on very-large instances}}
        To further evaluate the scalability of the proposed hybrid mechanism, we conducted experiments using the $\mathbb{B}$ instances \citep{ARNOLD201932}. In this set of experiments, the hybrid mechanism was applied exclusively to FILO2, as FILO2 is here recognized as a state-of-the-art metaheuristic for solving the very large-scale CVRP problems \citep{accorsi2021fast,ACCORSI2024106562}. \Cref{table:detailed-filo2-b} presents a comparative analysis between the original FILO2 algorithm and its hybrid variant, FILO2$^{*}$, on the $\mathbb{B}$ benchmark instances. 
        
\begin{figure}[htbp]  
    \centering
    \includegraphics[width=1\textwidth]{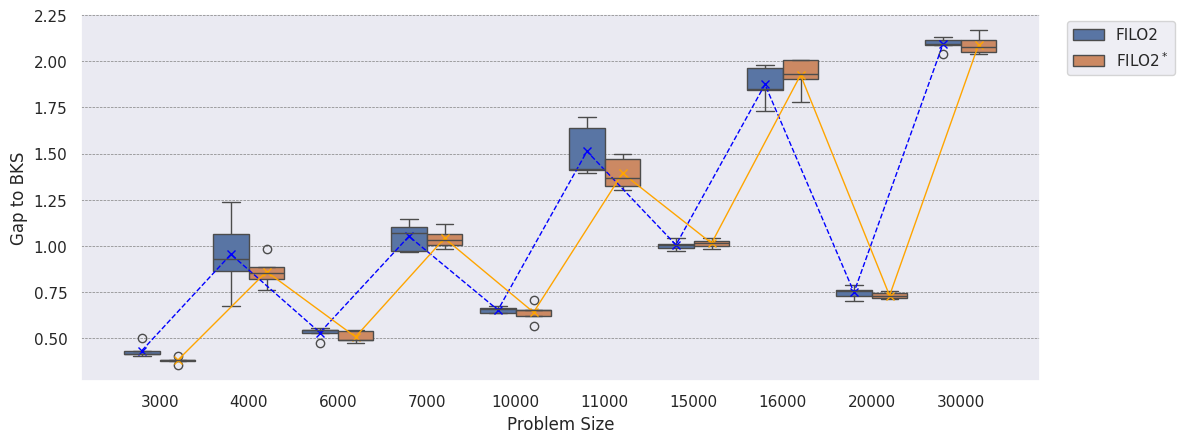}
    \caption{Comparison performance of FILO2 on $\mathbb{B}$ instances.} \label{fig:ch5-filo2-b} 
\end{figure}

        The results show that FILO2$^{*}$ generally achieves slightly better or at least comparable average gaps compared to the baseline FILO2, with consistent performance across nearly all instances. However, this improvement comes at the cost of slightly increased computation time, primarily due to the overhead of model inference using the deep learning-based selector $f_{\theta}$. Despite this additional cost, the average gap is reduced from $1.086\%$ to $1.059\%$, demonstrating an improvement from integrating the hybrid selector $f_{\theta}$ to the baseline algorithm within the core optimization phase. The performance trend is illustrated in \Cref{fig:ch5-filo2-b}.

\begin{table}[htbp] 
    \begin{center}
        \caption{Detailed result of FILO2 on $\mathbb{B}$ instances.}
        \label{table:detailed-filo2-b}
        \vspace*{0.2cm}
        \setlength\tabcolsep{2.2pt}
        \scalebox{0.8}{
            \begin{tabular}{l rr rrr r rrr r}
                \toprule
                    \multirow{2}{*}{\textbf{Instance}} 
                    &&  \multicolumn{3}{c}{\textbf{FILO2}} 
                    &&  \multicolumn{3}{c}{\textbf{FILO2$^{*}$}} 
                    &&  \multirow{2}{*}{\textbf{BKS}} \\
                    \cmidrule(lr){3-5} \cmidrule(lr){7-9}
                    &&  \textbf{Avg (Gap)} & \textbf{Best (Gap)} & \textbf{Time (s)} 
                    &&  \textbf{Avg (Gap)} & \textbf{Best (Gap)} & \textbf{Time (s)} 
                    &&\\
                \midrule
                    Leuven1 && 193683 (0.433) & 193627 (0.404) & 90.4 && \textbf{193683 (0.379)} & 193537 (0.357) & 112.8 && 192848\\
                    Leuven2 && 112454.2 (0.954) & 112145 (0.677) & 121 && \textbf{112454.2 (0.861)} & 112238 (0.76) & 154.6 && 111391\\
                    Antwerp1 && 479808.4 (0.53) & 479551 (0.476) & 98.4 && \textbf{479808.4 (0.507)} & 479538 (0.474) & 125.8 && 477277\\
                    Antwerp2 && 294414.8 (1.052) & 294173 (0.969) & 104.4 && \textbf{294414.8 (1.041)} & 294218 (0.984) & 137.2 && 291350\\
                    Ghent1 && 472596 (0.653) & 472507 (0.634) & 105.4 && \textbf{472596 (0.640)} & 472196 (0.568) & 137.4 && 469531\\
                    Ghent2 && 261645.2 (1.512) & 261346 (1.396) & 115.2 && \textbf{261645.2 (1.393)} & 261109 (1.304) & 147.6 && 257748\\
                    Brussels1 && \textbf{506761.8 (1.005)} & 506609 (0.975) & 102.6 && 506761.8 (1.014) & 506645 (0.982) & 142 && 501719\\
                    Brussels2 && \textbf{351955.2 (1.878)} & 351461 (1.735) & 113.4 && 351955.2 (1.929) & 351625 (1.782) & 159 && 345468\\
                    Flanders1 && 7294292 (0.748) & 7291040 (0.703) & 134.4 && \textbf{7294292 (0.731)} & 7291520 (0.71) & 189.6 && 7240118\\
                    Flanders2 && 4464854 (2.095) & 4462590 (2.043) & 144.4 && \textbf{4464854 (2.091)} & 4462450 (2.04) & 209.2 && 4373244\\
                \midrule
                    Average Gap && 1.086 & & 112.96 && \textbf{1.059} & & 151.52 && \\
                \bottomrule
            \end{tabular}
        }
    \end{center}
\end{table}

        In \Cref{fig:ch5-filo2-b} for baseline FILO2, on $\mathbb{B}$ instances, the one-tailed Wilcoxon signed-rank test rejects \Cref{hyp:ch5-a} with $p$-value $= 0.032$ $(< 0.05)$, indicating that proposed hybrid node destroyer statistically improve the performance of the baseline FILO2 when solving $\mathbb{B}$ instances. 
        
\begin{figure}[htbp]
    \centering
    \includegraphics[width=0.9\textwidth]{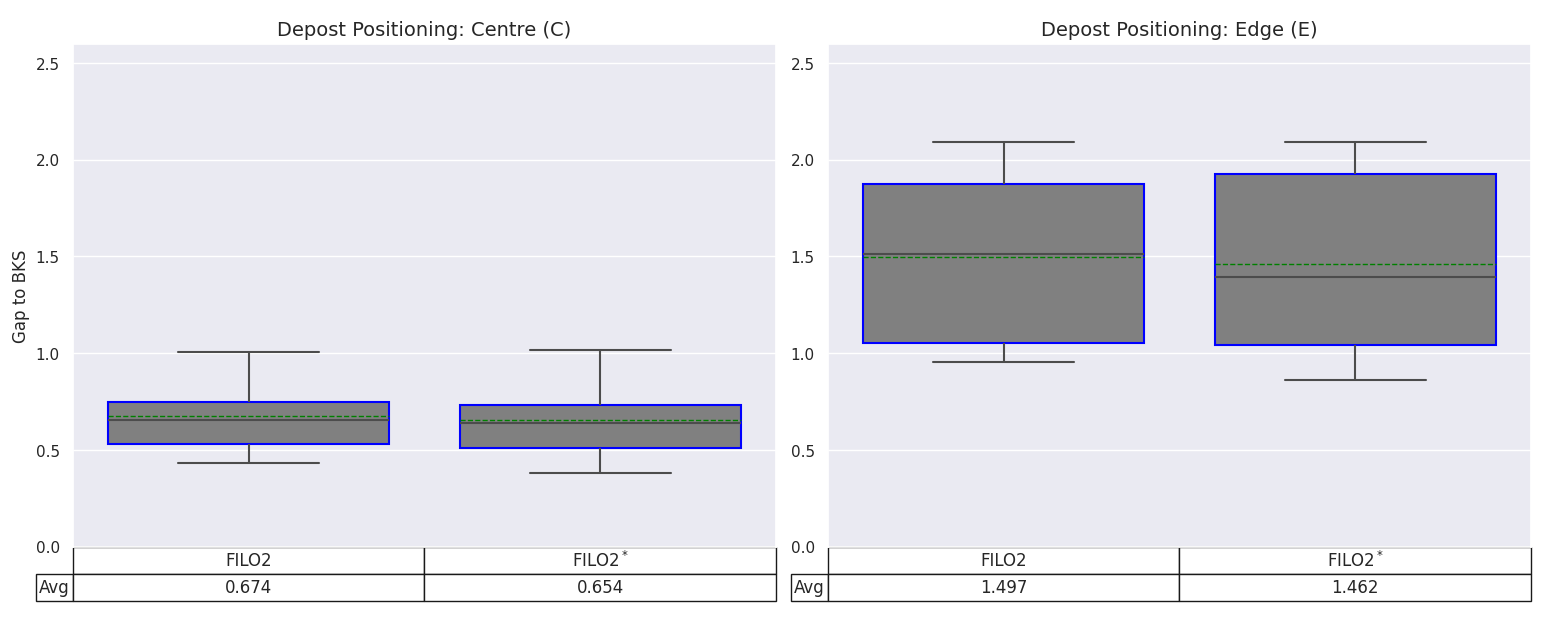}
    \caption{Performance according to depot positioning on $\mathbb{B}$ instances.} \label{fig:ch5-filo2-b-depot-boxplot} 
\end{figure}

        For further insight, \Cref{fig:ch5-filo2-b-depot-boxplot} provides a comparative boxplot illustrating the performance of FILO2 and FILO2$^{*}$ under different depot positioning scenarios. In the central (C) depot configuration, both FILO2 and FILO2$^{*}$ show low and consistent gap values, with FILO2$^{*}$ slightly outperforming the baseline. The tight distribution indicates stable performance when the depot is centrally located. In contrast, the edge (E) depot positioning scenario reveals significantly greater variability and higher average gap values. Nonetheless, FILO2$^{*}$ still achieves a marginal improvement, with an average gap of $1.462\% $compared to $1.497\%$ for the baseline. These results suggest that although FILO2 struggles more with edge depot configurations, the hybrid mechanism continues to offer performance gains regardless of depot positioning in the $\mathbb{B}$ instances.


\section{Conclusion} 
\label{sec:ch5-conclusion}
In this research, we proposed a novel iterative learning hybrid optimization framework to enhance the performance of metaheuristic solvers for the CVRP. Our approach integrates the Node Destroyer Model via selector $f_{\theta}$ into the metaheuristic framework designed to guide the LNS operators. This model learns to identify and preserve structurally significant customer nodes during the destroy phase of LNS. By excluding these nodes from modification, the algorithm maintains essential components of solutions, allowing for effective exploration. To develop this mechanism, we trained the selector model $f_{\theta}$ using a dataset of CVRP instances of size $N = 100$, and to address scalability, we applied a curriculum learning strategy and fine-tuned the model on instances with $N = 1,000$. The selector $f_{\theta}$ is subsequently embedded into two baselines: HGS-PILS and FILO2. We validated the effectiveness of the proposed approach through computational experiments. 

On $\mathbb{X}$ benchmark instances, our results show that the hybrid HGS-PILS$^{*}$ improves upon the performance of the baseline HGS-PILS. The improvements were also observed across various instance sizes and depot configurations. Furthermore, we assessed the scalability of our hybrid optimization strategy using the $\mathbb{B}$ instances, which represent very large-scale CVRP problems. FILO2 was chosen as the baseline due to its strong performance on such large-scale problems. The hybrid version, FILO2$^{*}$, demonstrated consistent improvements in the average gap over the baseline FILO2. Additionally, the hybrid mechanism maintained or improved performance across various depot positioning scenarios, further confirming the robustness of the approach. 

Overall, this study demonstrates the potential of integrating deep learning models with baseline metaheuristics to improve solution quality. By learning from high-quality solution structures, the hybrid model can guide the optimization process more intelligently than baseline heuristic operators. 

\paragraph{\textit{Limitations and future work}} 
    While the results highlight the effectiveness of the guided variant HGS-PILS$^{*}$ across a range of $\mathbb{X}$ instances and FILO2$^{*}$ on the large-scale $\mathbb{B}$ instances, several limitations merit further attention. Several directions offer potential for future research. One promising avenue is refining the guidance model by developing more generalizable hyperparameter settings that can be applied across a broader range of problems. Further improvements could also be achieved by incorporating more advanced learning architectures or by hybridizing the current selector $f_{\theta}$ with multiple heuristics to further enhance generalizability across diverse domains. 

\clearpage
\printbibliography

\end{document}